\documentclass[sigconf,authorversion]{acmart}

\usepackage{hyperref}
\usepackage{graphicx}
\usepackage{soul}
\usepackage{amsmath}
\usepackage{booktabs}
\usepackage{stmaryrd}
\usepackage{amsfonts,bm}
\usepackage{multirow}
\usepackage{cleveref}
\usepackage{enumitem}
\usepackage{balance}

\crefformat{section}{\S#2#1#3} 
\crefformat{subsection}{\S#2#1#3}
\crefformat{subsubsection}{\S#2#1#3}
\AtBeginDocument{%
  \providecommand\BibTeX{{%
    \normalfont B\kern-0.5em{\scshape i\kern-0.25em b}\kern-0.8em\TeX}}}

\setcopyright{acmcopyright}
\copyrightyear{2018}
\acmYear{2018}
\acmDOI{XXXXXXX.XXXXXXX}

\acmConference[Conference acronym 'XX]{Make sure to enter the correct
  conference title from your rights confirmation emai}{June 03--05,
  2018}{Woodstock, NY}
\acmPrice{15.00}
\acmISBN{978-1-4503-XXXX-X/18/06}

\newcommand{\specialcell}[2][c]{\begin{tabular}[#1]{@{}c@{}}#2\end{tabular}}
\newcommand\eg[0]{\textit{e.g.}}
\newcommand\ie[0]{\textit{i.e.}}

\begin{document}

\title{A Retrieve-and-Read Framework for Knowledge Graph Link Prediction}


\author{Vardaan Pahuja}
\email{pahuja.9@osu.edu}
\affiliation{%
  \institution{The Ohio State University}
  \city{Columbus}
  \state{Ohio}
  \country{United States}
}

\author{Boshi Wang}
\email{wang.13930@osu.edu}
\affiliation{%
  \institution{The Ohio State University}
  \city{Columbus}
  \state{Ohio}
  \country{United States}
}

\author{Hugo Latapie}
\email{hlatapie@cisco.com}
\affiliation{%
  \institution{Cisco Research}
  \city{San Jose}
  \state{California}
  \country{United States}
}

\author{Jayanth Srinivasa}
\email{jasriniv@cisco.com}
\affiliation{%
  \institution{Cisco Research}
  \city{San Jose}
  \state{California}
  \country{United States}
}

\author{Yu Su}
\email{su.809@osu.edu}
\affiliation{%
  \institution{The Ohio State University}
  \city{Columbus}
  \state{Ohio}
  \country{United States}
}


\begin{abstract}
Knowledge graph (KG) link prediction aims to infer new facts based on existing facts in the KG\@. Recent studies have shown that using the graph neighborhood of a node via graph neural networks (GNNs) provides more useful information compared to just using the query information. 
Conventional GNNs for KG link prediction follow the standard message-passing paradigm on the entire KG, which leads to superfluous computation, over-smoothing of node representations, and also limits their expressive power. On a large scale, it becomes computationally expensive to aggregate useful information from the entire KG for inference. To address the limitations of existing KG link prediction frameworks, we propose a novel retrieve-and-read framework, which first retrieves a relevant subgraph context for the query and then jointly reasons over the context and the query with a high-capacity reader.
As part of our exemplar instantiation for the new framework, we propose a novel Transformer-based GNN as the reader, which incorporates graph-based attention structure and cross-attention between query and context for deep fusion. This simple yet effective design enables the model to focus on salient context information relevant to the query. Empirical results on two standard KG link prediction datasets demonstrate the competitive performance of the proposed method. Furthermore, our analysis yields valuable insights for designing improved retrievers within the framework.\footnote{Code and data will be released on \url{https://github.com/OSU-NLP-Group/KG-R3/.}}
\end{abstract}





\maketitle
\renewcommand{\shortauthors}{Vardaan Pahuja, Boshi Wang, Hugo Latapie, Jayanth Srinivasa, \& Yu Su}

\section{Introduction}
Knowledge graphs encode a wealth of structured information in the form of \textit{(subject, relation, object)} triples. The rapid growth of KGs in recent years has led to their wide use in diverse applications such as information retrieval \citep{4039288, Zhang2016XKnowSearchEK}, question answering \citep{talmor-berant-2018-web, DBLP:conf/wsdm/HeL0ZW21}, and data mining \citep{zheng2021pharmkg}.
KG link prediction \citep{bordes2013translating} that aims to infer new facts based on existing facts is a fundamental task on KGs. It finds applications in relation extraction \citep{wang2014knowledge, chen-badlani-2020-relation}, question answering \citep{10.1145/3289600.3290956, saxena-etal-2020-improving}, and recommender systems \citep{zhang2016collaborative, 9702632}.

Early methods for KG link prediction have focused on learning a dense embedding for each entity and relation in the KG, which is then used to calculate the plausibility of new facts via a simple scoring function \citep{bordes2013translating, socher2013reasoning, lin2015learning, ji2015knowledge, dettmers2018convolutional}. 
The hope is that an entity's embedding will learn to compactly encode the structural and semantic information such that a simple scoring function would suffice for making accurate link predictions.
However, it is challenging to fully encode the rich information of KGs into such shallow embeddings.
Similar to how contextualized encoding models like BERT \citep{devlin-etal-2019-bert} have been supplanting static embeddings \citep{mikolov2013distributed, pennington2014glove} for natural language representation, several recent studies have adapted message-passing graph neural networks (GNNs) for KG link prediction \citep{schlichtkrull2018modeling, shang2019end, vashishth2020compositionbased}.
By using a GNN to iteratively encode increasingly larger graph neighborhoods, GNN-based KG link prediction methods have shown great success.
However, the reliance on message-passing over the entire KG leads to superfluous computation and limits their scalability on large-scale KGs.
The same reason also leads to slow inference speed.

\begin{figure*}[!ht]
  \centering
  \includegraphics[width=\linewidth]{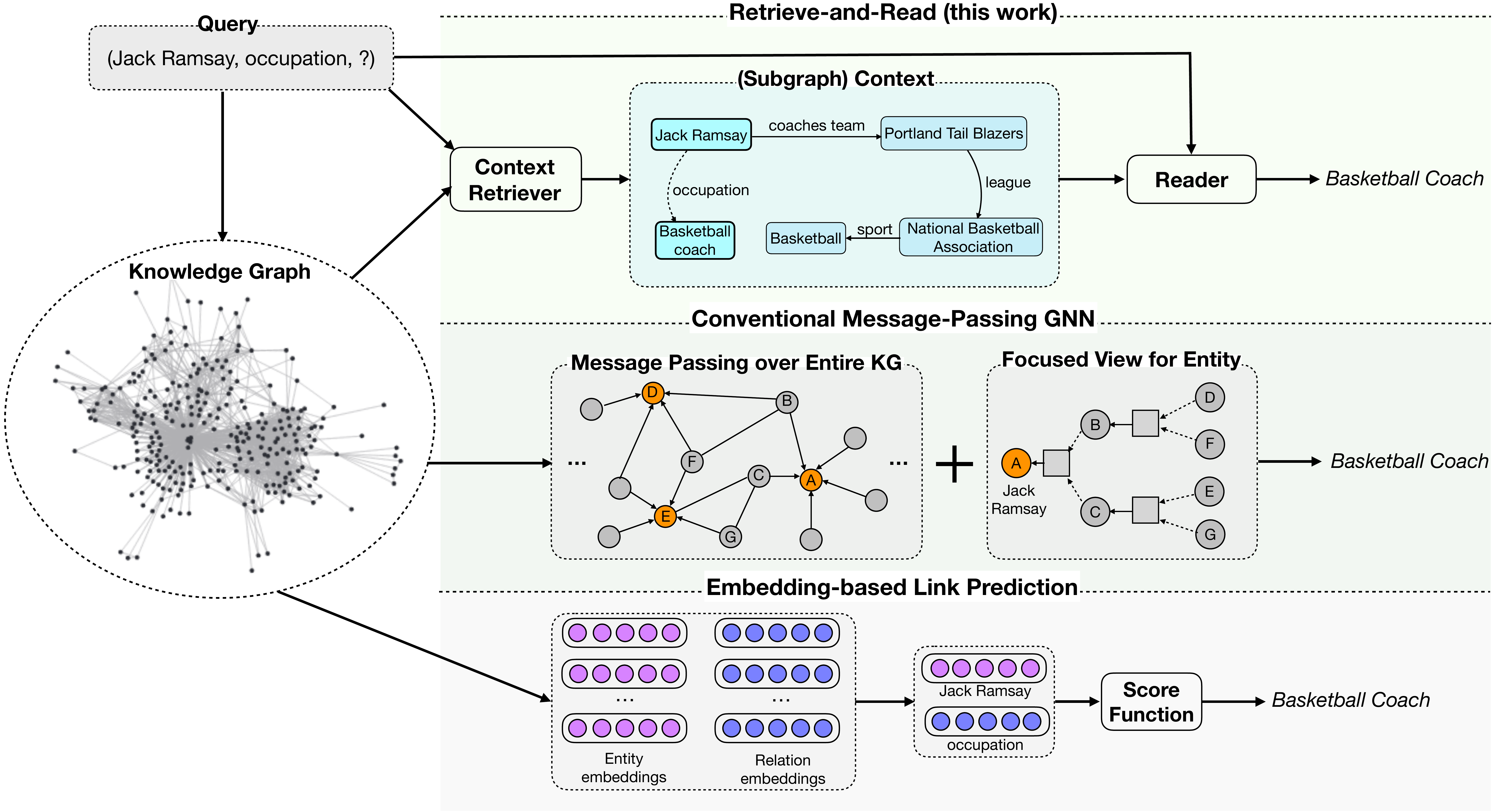}
  \caption{Overview of the proposed retrieve-and-read framework and comparison with existing frameworks for KG link prediction. Embedding-based methods try to encode all relevant information into shallow embeddings, while message-passing graph neural networks (GNNs) iteratively learn the representations through message-passing over the entire KG\@. In contrast, in our framework, we first retrieve a small context subgraph that is relevant to each input query, and then jointly encode the query and the context for the final prediction. Here for simplicity, we assume the context to be a connected subgraph, but being connected is not a necessary condition.}
  \label{fig:hook_fig}
\end{figure*}

Intuitively, for each specific query, \eg, \textit{(Jack Ramsay, occupation, ?)}, only a small subgraph of the entire KG may be relevant for answering the query (Figure~\ref{fig:hook_fig}). If we could \textit{retrieve} the relevant subgraph from the KG as \textit{context}, we can then easily use a high-capacity model to \textit{read} the query in the corresponding context to make the final inference. 
To this end, we propose a novel \textit{retrieve-and-read framework} for KG link prediction (see Figure~\ref{fig:hook_fig} for an overview and comparison with existing frameworks). 
It consists of two main components: a \textit{retriever} that identifies the relevant KG subgraph as context for the input query and a \textit{reader} that jointly considers the query and the retrieved context for inferring the answer.
Such a retrieve-and-read framework has been widely used for the open-domain question answering (QA) problem \citep{chen-etal-2017-reading, zhu2021retrieving}, which faces a similar fundamental challenge: it also needs to answer a question in a massive corpus where only a small fraction is relevant to each specific question.
The modularization provided by this framework has enabled rapid progress on retriever and reader models separately \citep{karpukhin-etal-2020-dense, glass-etal-2020-span, xiong2021answering, khattab2021baleen}.

Embracing the retrieve-and-read framework for KG link prediction could bring multiple potential advantages: 
1) It provides great flexibility to explore and develop diverse models for the retriever and reader separately. For example, we explore several different choices for the retriever, even including an existing KG link prediction model. Because the reader only needs to deal with a small subgraph instead of the entire KG, we could leverage the potential of high-capacity and more expressive models such as the Transformer \citep{vaswani2017attention}, which has proven extremely successful for other tasks, but the application on KG link prediction has been limited. 
2) Relatedly, separate and more focused progress can be made on each component, which can then be combined to form new KG link prediction models. 
3) Instead of learning and leveraging the same \textit{static} representation for all inferences as in existing frameworks, the reader can dynamically learn a contextualized representation for each query and context for more accurate prediction. 
4) Finally, this framework could potentially support the development of GNN-based KG link prediction models for large-scale KGs, similar to how it has enabled open-domain QA to operate on web-scale corpora \citep{chen-etal-2017-reading, DBLP:conf/aaai/WangYGWKZCTZJ18, karpukhin-etal-2020-dense, zhu2021retrieving}.

To demonstrate the effectiveness of the proposed framework, we propose a novel instantiation of the framework, KG-R3 (\underline{KG} \underline{R}easoning with \underline{R}etriever and \underline{R}eader). 
It uses an existing KG link prediction method, MINERVA \citep{das2018go} as retriever and a novel \textit{Transformer-based GNN} as the reader.
Existing GNNs are mostly based on message-passing \citep{gilmer2017neural}, where the representation of a particular node is iteratively updated by its neighbors. 
Message-passing GNNs suffer from limited expressive power \citep{DBLP:conf/iclr/XuHLJ19} and over-smoothing \citep{li2018deeper}, \ie, the representation of distinct nodes become indistinguishable as GNNs get deeper, which limits their model capacity.
The time complexity of message-passing also grows exponentially with the number of layers, making it even harder to increase the capacity of such models. 
On the other hand, the Transformer model \citep{vaswani2017attention} has been driving the explosive growth of high-capacity models such as BERT \citep{devlin-etal-2019-bert}. 
The Transformer can support very high-capacity models \citep{NEURIPS2020_1457c0d6, chowdhery2022palm}, which is one of the key reasons for its success.
While it is challenging to apply Transformer to the entire KG due to its limited context window, the small context in the retrieve-and-read framework makes it feasible. 
We design a novel Transformer-based GNN which has a two-tower structure to separately encode the query and the context subgraph and a cross-attention mechanism to enable deep fusing of the two towers. 
A graph-induced attention structure is also developed to encode the context subgraph.

The major contribution of this work is three-fold:
\begin{itemize}
    \item We propose a novel retrieve-and-read framework for knowledge graph link prediction.
    \item We develop a novel instantiation, KG-R3, of the framework, which consists of a novel Transformer-based graph neural network for KG link prediction.
    \item We conduct empirical experiments on the standard FB15K-237 \citep{toutanova2015observed} and WN18RR \citep{dettmers2018convolutional} datasets and show that KG-R3 achieves competitive results with state-of-the-art methods.
\end{itemize}

\section{Related Work}
\textbf{Graph Neural Networks.} 
Graphs naturally encode rich semantics of underlying data. 
Early models \citep{bruna2013spectral, defferrard2016convolutional, kipf2017semisupervised} extended the spectral convolution operation to graphs. Follow-up work \citep{bresson2017residual, velickovic2018graph, DBLP:conf/wsdm/JinDW0LT21} introduced attention and gating mechanisms to aggregate the salient information from a node's neighborhood.
These aforementioned models are applicable only to homogeneous graphs.
In our present work, we develop a novel Transformer-based GNN as the reader module for link prediction in multi-relational graphs like KGs.\\

\noindent \textbf{KG Link Prediction Models.}
Early approaches for link prediction range from translation-based models \citep{bordes2013translating, lin2015learning} and semantic matching models \citep{nickel2011three, yang2014embedding} to the ones that leverage CNNs \citep{dettmers2018convolutional, nguyen2017novel}. These shallow embedding methods learn embeddings for each entity and relation and use a parameterized score function to predict the plausibility of a triple.
To make use of the semantically rich graph neighborhood, several approaches have tried to adapt GNNs to multi-relational graphs for KG link prediction.
R-GCN \citep{schlichtkrull2018modeling} and CompGCN \citep{vashishth2020compositionbased} make use of relation-type dependent message aggregation. 
These methods use graph aggregation over the entire KG, thus limiting their scalability.
In our framework, we propose to perform computation only on a relevant subgraph of the KG to mitigate this problem.\\

\noindent \textbf{Path-based KG Reasoning.}
Another line of work uses multi-hop paths to synthesize information for predicting the missing facts in a KG\@. DeepPath \citep{xiong-etal-2017-deeppath} and MINERVA \citep{das2018go} formulate it as a sequential decision-making problem and use reinforcement learning to search paths to the target entity.
For the retriever module, we use MINERVA as one of the baseline methods. Neural LP \citep{yang2017differentiable} and DRUM \citep{sadeghian2019drum} use inductive logic programming to learn logical rules from KGs, which are used to weight different paths.
Though these approaches are interpretable, they suffer from relatively poor performance compared to embedding-based KG link prediction methods.
Our proposed framework can utilize the subgraphs generated by these approaches for improved performance.\\

\noindent \textbf{Transformers for Graph ML Tasks.}
In contrast to NLP and vision, developing transformer models for graphs is challenging due to the absence of regularity in the data space, for example, 2D/3D grid in images and linear structure in sentences.
\citet{DBLP:conf/iclr/VelickovicCCRLB18, dwivedi2021generalization} use Transformer-like self-attention for neighborhood aggregation in message-passing GNNs. \citet{dwivedi2020benchmarking, kreuzer2021rethinking, kim2022pure} introduce positional encodings (PE) as an inductive bias for graph structure. Heterogeneous Graph Transformer \citep{yao2020heterogeneous} proposes a Levi graph based attention structure for encoding Abstract Meaning Representation (AMR) graphs for the task of AMR-to-text generation.
GRAPH-BERT \citep{zhang2020graph} uses the Transformer model for self-supervised learning of node representations, which can then be fine-tuned on downstream tasks.
Similarly, Transformers have been applied for the task of KG link prediction. HittER \citep{chen-etal-2021-hitter} proposes a hierarchical Transformer model, which utilizes one-hop neighborhood context for the KG link prediction task. StAR \citep{DBLP:conf/www/WangSLZW021} augments the contextual text representation obtained from a pre-trained language model with the structure information to learn better representations. Similarly, kgTransformer \citep{DBLP:conf/kdd/LiuZSCQZ0DT22} pre-trains a Transformer model for out-of-domain generalization to unseen first-order logic queries.
This work focuses on developing a Transformer-based GNN as the reader that can handle arbitrary subgraphs as context.\\

\noindent \textbf{Open-domain Question Answering.}
The task of open-domain QA is to answer a question using knowledge from a massive corpus such as Wikipedia. A popular and successful way to address the challenge of large scale is through a two-stage retrieve-and-read pipeline \citep{chen-etal-2017-reading, zhu2021retrieving}, which leads to rapid developments of retriever and reader separately \citep{karpukhin-etal-2020-dense, xiong2021answering, khattab2021baleen, deng-etal-2021-reasonbert}. We draw inspiration from this pipeline and propose to use a retrieve-and-read framework for KG link prediction.

\section{Methodology}
\subsection{Preliminaries}
\noindent \textbf{Knowledge Graph.} Given a set of entities $\mathcal{E}$ and a set of relations $\mathcal{R}$, a knowledge graph can be defined as a collection of facts $\mathcal{F} \subseteq \mathcal{E} \times \mathcal{R} \times \mathcal{E}$ where for each fact $f = (h, r, t)$, $\;h,t \in \mathcal{E}, r \in \mathcal{R}$.\\

\noindent \textbf{Link Prediction.} The task of link prediction is to infer the missing facts in a KG\@. Given a link prediction query $(h, r, ?)$ or $(?, r, t)$, the model ranks the target entity among the set of candidate entities. For the query $(h, r, ?)$, $h$ and $r$ correspond to the source entity and the query relation, respectively.
\subsection{Retriever}
The function of the retriever module is to select a relevant subgraph of the KG as the query context.
We use the following off-the-shelf methods to generate subgraph inputs for the Transformer-based reader module in our framework.

\begin{figure*}[htbp]
  \centering
  \includegraphics[width=\textwidth]{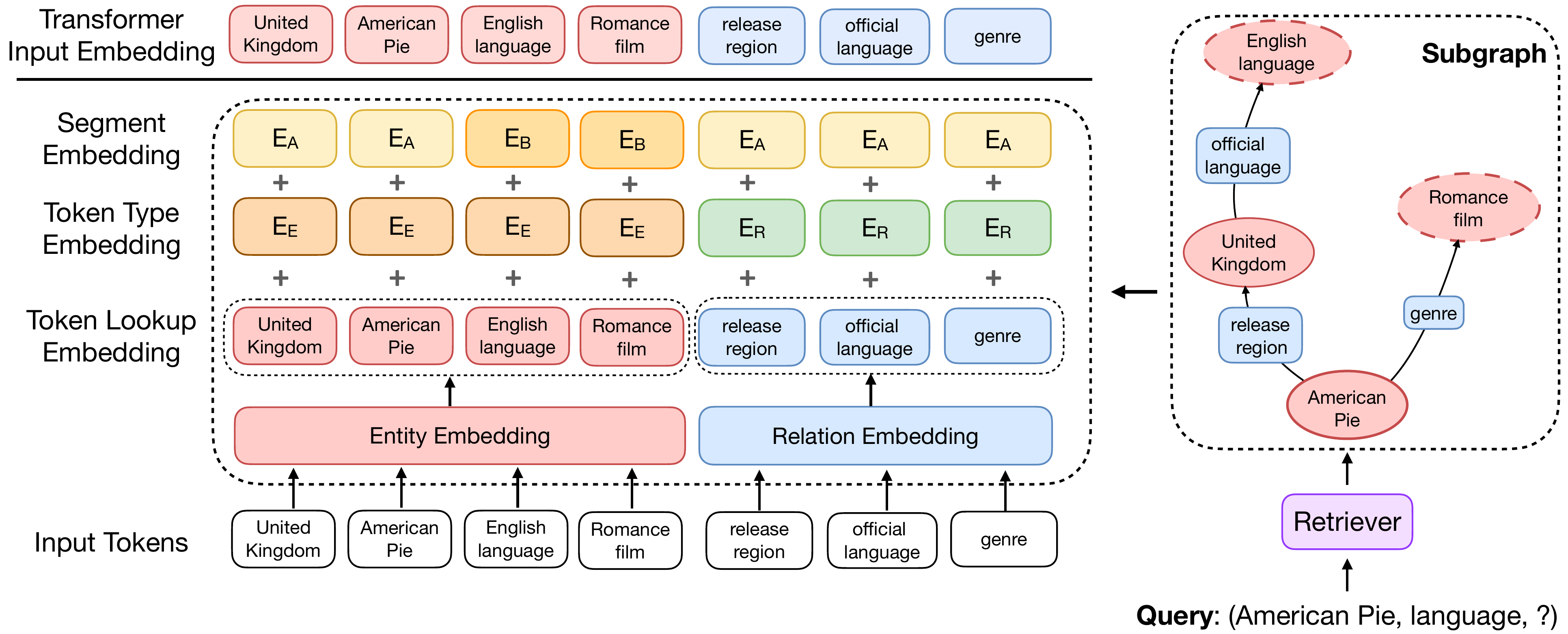}
  \caption{Schematic representation of embedding layer for subgraph input. The Transformer input is the sum of the token lookup embedding, the token type embedding, and the segment embedding.}
  \label{fig:model_embed}
\end{figure*}

\subsubsection{Uninformed Search-based Retrieval Strategies}
\begin{enumerate}[label=\alph*)]
\item\textbf{Breadth-first search}: For breadth-first search, we sample edges starting from the source entity in breadth-first order till we reach the context budget.
\item\textbf{One-hop neighborhood}: The one-hop neighborhood includes edges in the immediate one-hop neighborhood of the source entity.
\end{enumerate}

\subsubsection{Informed Search-based Retrieval Strategies}

We use \textbf{MINERVA} \citep{das2018go} as our informed search-based retrieval strategy.
It formulates KG link prediction as the generation of multi-hop paths from the source entity to the target entity. The environment is represented as a Markov Decision Process (MDP) on the KG, where the reinforcement learning agent gets a positive reward on reaching the target. The set of paths generated by MINERVA provides an interpretable provenance for KG link prediction. The retriever model utilizes the union of these paths decoded using beam search as the subgraph output.

Among these approaches, breadth-first search and one-hop neighborhood use uninformed search, \ie, they only enrich the query using surrounding context without explicitly going toward the target entity. On the other hand, the subgraph obtained using the MINERVA retriever aims to provide context which encloses information towards reaching the target entity.

\subsection{Reader Architecture}

\noindent \textbf{Embedding Layer.}
The input to the Transformer is obtained by summing the token lookup embedding, the token type embedding, and the segment embedding (Figure~\ref{fig:model_embed}):
\begin{itemize}
    \item \textbf{Token lookup embedding}: We use learned lookup embeddings for all entities and relations. These lookup embeddings store the global semantic information for each token.
    \item \textbf{Token type embedding}: Entity and relation tokens have different semantics, so we use token type embeddings to help the model distinguish between them.
    \item \textbf{Segment embedding}: It denotes whether a particular entity token corresponds to the terminal entity in a path beginning from the source entity. This helps the model to differentiate between the terminal tokens, which are more likely to correspond to the final answer vs.\ other tokens.
\end{itemize}
The input to the reader module is the query, \eg, $(h, r, ?)$, and its associated context, a subgraph of the KG\@. The input sequence for the subgraph encoder is constructed by concatenating the sets of node and edge tokens.
Each edge corresponds to a unique token in the input, though there might be multiple edges with the same predicate.
At a high level, the query and subgraph are first encoded by their respective Transformer encoders.  The query self-attention encoder takes ``\texttt{[CLS], [source entity], [query relation]}'' as the input sequence with a fully-connected attention structure. Then the cross-attention module is used to modulate the subgraph representation, conditioned on the query.\\

\begin{figure}[htbp]
    \centering
    \includegraphics[width=0.45\textwidth]{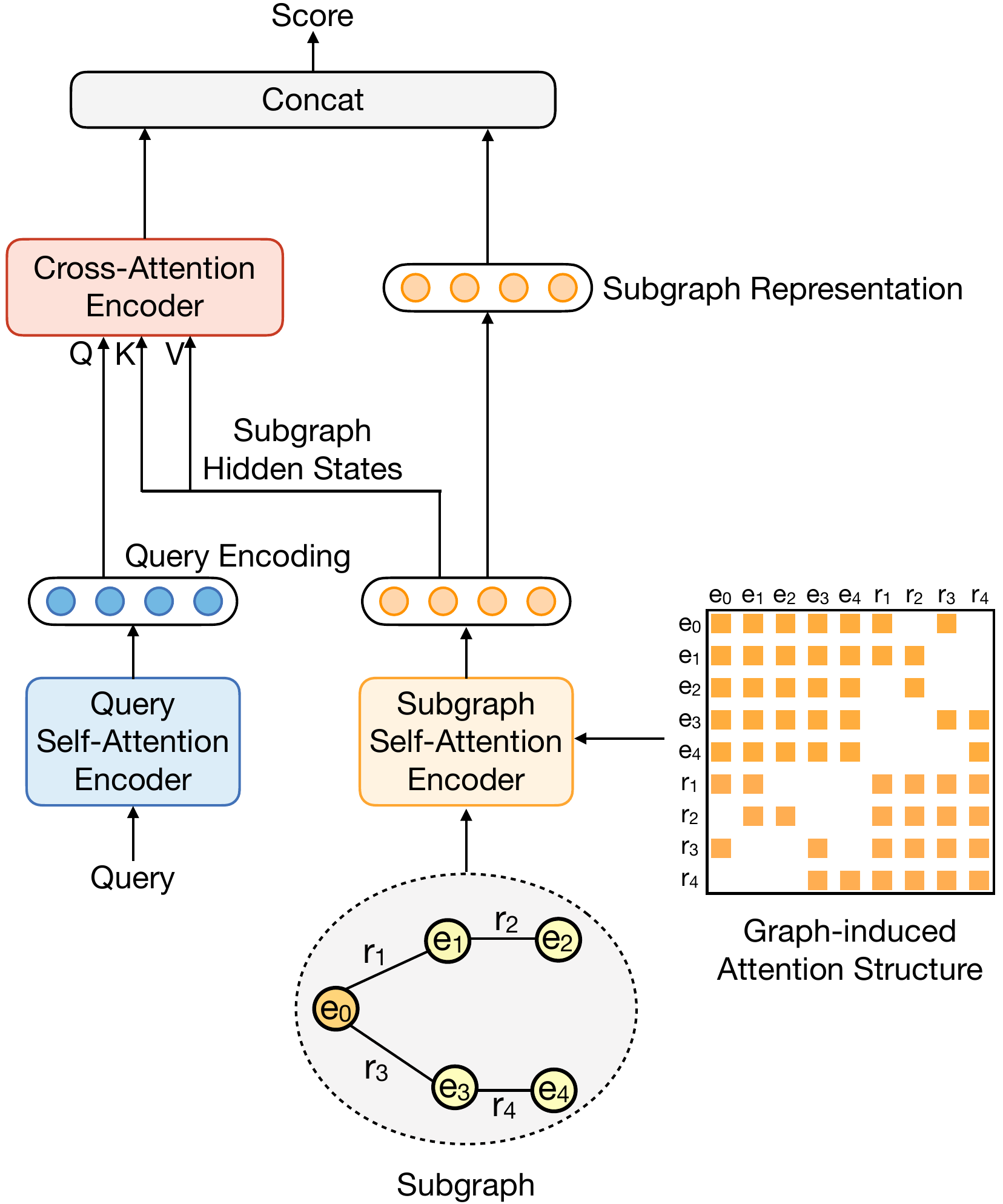}
    \caption{Reader module architecture. In the subgraph self-attention encoder, the graph-induced self-attention design governs the attention between tokens. In the cross-attention encoder, the link prediction query hidden states serve as the query input, while the subgraph hidden states serve as the key and value inputs.}
    \label{fig:model_overall}
\end{figure}

\noindent \textbf{Graph-induced Self-Attention.} The attention structure ($\mathcal{A}_i$) governs the set of tokens that a particular token can attend to in the self-attention layer of the subgraph encoder. It helps incorporate the (sub)graph structure into the transformer representations.
Inspired by KG-augmented PLMs \cite{he-etal-2020-bert, he2021klmo}, we define the attention structure (Figure~\ref{fig:model_overall}) such that 1) all node tokens can attend to each other; 2) all edge tokens can attend to each other; 3) for a particular triple $(h, r, t)$, the token pairs $(h, r)$ and $(r, t)$ can attend to each other, and 4) each token attends to itself. This design is motivated by the need to maintain a balance between the immediate graph neighborhood of a token vs.\ its global context in the subgraph.

More formally, let the subgraph consist of $m$ nodes and $n$ edges.
Let ${\{\bm{h}_i^{\ell}\}}_{i=1}^{m+n}$ denote the hidden representations of the tokens in layer $\ell$ of the subgraph self-attention encoder,
\begin{align}
    \bm{h}_{i}^{\ell+1} &= \mathbf{O}^{\ell} \ \bigparallel_{k=1}^{H} \bigg(\sum_{j \in \mathcal{A}_i} w_{ij}^{k,\ell} \mathbf{V}^{k,\ell}\bm{h}_j^{\ell} \bigg),\\
    w_{ij}^{k,\ell} &= \textnormal{softmax}_{j \in \mathcal{A}_i} \bigg( \frac{\mathbf{Q}^{k, \ell} \bm{h}_i^{\ell}\cdot \mathbf{K}^{k, \ell}\bm{h}_j^{\ell}}{\sqrt{d_k}}\bigg).
\end{align}
\noindent Here, $\mathbf{Q}^{k,\ell}, \mathbf{K}^{k,\ell}, \mathbf{V}^{k,\ell} \in \mathbb{R}^{d_k \times d}$, $\mathbf{O}^{\ell} \in \mathbb{R}^{d \times d}$ are projection matrices for the $k^{th}$ attention head in layer $l$, $H$ denotes the number of attention heads per layer, $d_k$ denotes the hidden dimension of keys and $\|$ denotes concatenation.\\

\noindent \textbf{Cross-Attention.} To answer a link prediction query, the model needs a way to filter the context in the subgraph that is relevant to a particular link prediction query. To accomplish this, we introduce cross-attention from the query to the subgraph (Figure~\ref{fig:model_overall}). Following \citet{vaswani2017attention}, the queries come from query hidden states, whereas the keys and values come from the subgraph hidden states. The resultant representation encodes the subgraph information that is relevant to the query at hand. 
More formally, Let $\bm{e}^{\textnormal{q}, \ell}_i$ and $\{\bm{e}^{\textnormal{sub}}_j\}_{j=1}^{m+n}$ denote the self-attention hidden representations of $i^{th}$ query token and the subgraph tokens, respectively,

\begin{align}
&\textrm{Cross-Attention}(\{\bm{e}^{\textnormal{q}, \ell}_i\}, \{\bm{e}^{\textnormal{sub}}_j\}_{j=1}^{m+n}) \nonumber\\
&= \mathbf{O}^{\ell} \ \bigparallel_{k=1}^{H} \bigg(\sum_{j=1}^{m+n} w_{ij}^{k,\ell} \mathbf{V}^{k,\ell}\bm{e}_j^{\textnormal{sub}} \bigg),\\
w_{ij}^{k,\ell} &= \textnormal{softmax}\bigg( \frac{\mathbf{Q}^{k, \ell} \bm{e}_i^{\textnormal{q}, \ell}\cdot \mathbf{K}^{k, \ell}\bm{e}_j^{\textnormal{sub}}}{\sqrt{d_k}}\bigg).
\end{align}

\noindent This is concatenated with the contextualized representation of the source entity in the subgraph to output the feature vector for predicting the plausibility scores.
For a link prediction query \eg, $(h, r, ?)$, the model predicts a score distribution over all candidate entities. The model is trained using cross-entropy loss, framing it as a multi-class classification problem. Figure~\ref{fig:model_overall} illustrates the overall model architecture of the Transformer-based reader.
\begin{table}[htbp]
\centering
\caption{Dataset Statistics}
\label{tab:dataset_stat}
\begin{tabular}{cccccc}
\toprule
\bfseries Dataset   & $\bm{|\mathcal{E}|}$      & \bfseries $\bm{|\mathcal{R}|}$   & \multicolumn{3}{c}{\# \bfseries Facts} \\ \cmidrule{4-6}
          &        &     & \bfseries Train    & \bfseries Valid   & \bfseries Test                      \\ \midrule
FB15K-237 & 14,541 & 237 & 272,115  & 17,535  & 20,466            \\
WN18RR    & 40,943 & 11  & 86,835   & 3,034   & 3,134       \\ \bottomrule        
\end{tabular}
\end{table}
\begin{table*}[htbp]
\centering
\caption{Comparison of our framework with baseline methods. For all metrics, higher is better. Missing values are denoted by --. Results of RESCAL, TransE, DistMult, ComplEx, and ConvE correspond to the best results obtained after extensive hyperparameter tuning \citep{Ruffinelli2020You}.
Result of Neural LP and DRUM are taken from \citet{DBLP:conf/iclr/QuCXBT21} following the standard evaluation setting.
Results of other methods are taken from their original papers.
}
\label{tab:results_overall}
\begin{tabular}{llcccc|cccc}
\toprule
& & \multicolumn{4}{c}{\bfseries FB15K-237} & \multicolumn{4}{c}{\bfseries WN18RR} \\ \cmidrule{2-10}
\bfseries Framework & \bfseries Model                                  & \bfseries MRR        & \bfseries Hits@1    & \bfseries Hits@3    & \bfseries Hits@10   & \bfseries MRR      & \bfseries Hits@1   & \bfseries Hits@3   & \bfseries Hits@10   \\ \midrule
\textbf{Embedding-} & RESCAL \citep{nickel2011three} & .356	& .263 & .393 & .541 & .467	& .439 & .480 & .517 \\
\textbf{based} & TransE \citep{bordes2013translating} & .313 & .221 & .347 & .497 & .228 & .053 & .368 & .520  \\
& DistMult \citep{yang2014embedding} & .343	& .250 & .378 & .531 & .452	& .413 & .466 & .530\\
& ComplEx \citep{trouillon2016complex} & .348 & .253 & .384 & .536 & .475 & .438 & .490 & .547\\
& ComplEx-N3 \citep{Lacroix2018CanonicalTD} & .37 & -- & -- & .56 & .48 & -- & -- & .57 \\
& RotatE \citep{sun2018rotate} & .338 & .241 & .375 & .533 & .476 & .428 & .492 & .571\\ \midrule
\textbf{CNN-based} & ConvKB \citep{nguyen2017novel} & .243 & .155 & .371 & .421 & .249 & .057 & .417 & .524 \\
& ConvE \citep{dettmers2018convolutional} & .339 & .248	& .369 & .521 & .442 & .411 & .451 & .504\\
\midrule
\textbf{Path-based} & MINERVA \citep{das2018go} & .293 & .217 & .329 & .456 & .448 & .413 & .456 & .513\\
& Neural LP \citep{yang2017differentiable} &  .237 & .173 & .259 & .361 & .381 & .368 & .386 & .408\\
& DRUM \citep{sadeghian2019drum} & .238 & .174 & .261 & .364 & .382 & .369 & .388 & .410\\
\midrule
\textbf{GNN-based} & R-GCN \citep{schlichtkrull2018modeling} & .248 & .151 & -- & .417 & -- & -- & -- & -- \\
& CompGCN \citep{vashishth2020compositionbased} & .355 & .264 & .390 & .535 & .479 & .443 & .494 & .546 \\
\midrule
\textbf{Transformer-based} & HittER \citep{chen-etal-2021-hitter} & .373 & .279 & .409 & \textbf{.558} & \textbf{.503} & \textbf{.462} & \textbf{.516} & \textbf{.584} \\
 \midrule
& \textbf{KG-R3} (this work) & \textbf{.390} & \textbf{.315}  & \textbf{.413}  & .539 & .472 & .439 & .481 & .537 \\ \bottomrule
\end{tabular}
\end{table*}

\section{Experiments} \label{sec:expt}

\subsection{Datasets}
We use standard link prediction benchmarks FB15K-237 \citep{toutanova2015observed} and WN18RR \citep{dettmers2018convolutional} to evaluate our model.
FB15K-237 is an encyclopedic KG derived from the FB15K dataset \citep{bordes2013translating} after removing data contamination due to overlap between train and test sets. In a similar fashion, WN18RR is derived from the WN18 dataset \citep{bordes2013translating} after deleting inverse relations to prevent inference by memorizing training facts. The FB15K-237 dataset is released under Microsoft Research Data License Agreement. The WN18RR dataset is released by Princeton University under their license. Both datasets are permitted to be used for academic research in their original form. 
Dataset statistics are reported in Table~\ref{tab:dataset_stat}.


\begin{table*}[htbp]
\centering
\caption{Comparison of our Transformer-based GNN with other reader architectures using MINERVA subgraphs. Please refer to the Appendix for implementation details.}
\label{tab:reader_baselines}
\begin{tabular}{ccccc|cccc}
\toprule
& \multicolumn{4}{c}{\bfseries FB15K-237} & \multicolumn{4}{c}{\bfseries WN18RR} \\ \cmidrule{1-9}
\bfseries Model                                  & \bfseries MRR        & \bfseries Hits@1    & \bfseries Hits@3    & \bfseries Hits@10   & \bfseries MRR      & \bfseries Hits@1   & \bfseries Hits@3   & \bfseries Hits@10   \\ \midrule
CompGCN \cite{vashishth2020compositionbased} & .272 & .185 & .302 & .444 & .335 & .267 & .382 & .452 \\
HetGT \cite{yao2020heterogeneous} & .264 & .213 & .274 & .361 & .365 & .317 & .396 & .443  \\ \midrule
\textbf{Ours} & \textbf{.390} & \textbf{.315}  & \textbf{.413}  & \textbf{.539} & \textbf{.472} & \textbf{.439} & \textbf{.481} & \textbf{.537} \\ \bottomrule
\end{tabular}
\end{table*}

\subsection{Evaluation Protocol}
For a link prediction query, \eg, $(h, r, ?)$, the model scores all candidate triplets $\mathcal{C} = \{(h, r, t^{'}), t^{'} \in \mathcal{E}\}$ and ranks the correct entity in the list.
Following \citet{bordes2013translating}, we use the filtered evaluation setting \ie, the rank of a target entity is not affected by alternate correct entities. We report results on standard evaluation metrics: Mean Reciprocal Rank (MRR), Hits@1, Hits@3, and Hits@10.

\subsection{Baselines}

We conduct a comprehensive comparison of our model with various KG link prediction baselines. These include embedding-based methods such as TransE \cite{bordes2013translating}, DistMult \cite{yang2014embedding}, and ComplEx \cite{trouillon2016complex}, CNN-based approaches like ConvE \cite{dettmers2018convolutional} and ConvKB \cite{nguyen2017novel}, path-based techniques such as MINERVA \cite{das2018go}, Neural LP \cite{yang2017differentiable}, and DRUM \citep{sadeghian2019drum}, GNN-based methods like R-GCN \cite{schlichtkrull2018modeling} and CompGCN \cite{vashishth2020compositionbased}, as well as transformer-based approaches like HittER \cite{chen-etal-2021-hitter}.\footnote{Due to space constraints, we only include the top performing KG link prediction methods in Table~\ref{tab:results_overall}. For comprehensive results on all approaches in the literature, please refer to the surveys \cite{nguyen-2020-survey, ali2021bringing}.} To ensure a fair comparison, we exclude models that involve excessive computation, such as NBFNet \cite{zhu2021neural}, which relies on learning representations based on \textit{all} paths between each pair of entities. This computationally expensive operation limits its scalability on large KGs. Additionally, we exclude approaches that utilize extra information or pre-trained language models \cite{DBLP:conf/www/WangSLZW021, lovelace-rose-2022-framework}.

\subsection{Implementation Details}
We implement our models in Pytorch.
We use $L=3, A=8, H=320$ for the Transformer model (both self-attention and cross-attention layers), where $L$, $A$, and $H$ denote the number of layers, the number of attention heads per layer, and the hidden size, respectively.
The intermediate hidden dim. for the feedforward layer in Transformer is set to 1280. 
We use the Adamax \citep{DBLP:journals/corr/KingmaB14} optimizer for training. The learning rate schedule includes warmup for $10\%$ of the training steps followed by linear decay.
The batch size is set to 512. For both datasets, we tune the learning rate on the validation set and report results on the test set with the best validation setting. 
We sort the training instances in ascending order based on subgraph size for better training efficiency. During training, we use early stopping based on the validation set to prevent overfitting. To prevent exposure bias, we omit subgraph edges that overlap with the query triple during training.
A more comprehensive description of the hyperparameter setup is given in the Appendix.

\section{Results}
In this section, we attempt to answer the following questions:
\begin{enumerate}
    \item [Q1.] How does the performance of KG-R3 compare to other KG link prediction baselines? (\cref{sec:main_results})
    \item [Q2.] How does our proposed Transformer-based GNN reader compare to other reader architectures? (\cref{sec:other_reader_comparison})
    \item [Q3.] How do factors like the presence of the target entity in the subgraph and the entity degree affect the link prediction performance? (\cref{sec:target_ent_present_analysis})
    \item [Q4.] How does KG-R3 compare to other GNN baselines in terms of theoretical complexity and inference speed? (\cref{sec:efficiency})
\end{enumerate}

\subsection{Performance Comparison on KG Link Prediction} \label{sec:main_results}
Table~\ref{tab:results_overall} shows the overall link prediction results. 
On the FB15K-237 benchmark, KG-R3 consistently outperforms all baselines in 3 out of 4 evaluation metrics.
Our model obtains a $12.9\%$ relative improvement (Hits@1) over HittER \citep{chen-etal-2021-hitter}, a baseline Transformer model for KG link prediction. Compared to the best performing message-passing GNN baseline (CompGCN), it improves by 5.1\% in terms of Hits@1.
Our model, which uses the MINERVA subgraph retriever, improves over the MINERVA baseline by 9.8\% and 2.6\% (Hits@1) for FB15K-237 and WN18RR, respectively.
For the WN18RR dataset, our model demonstrates superior performance compared to embedding-based approaches, CNN-based approaches, and path-based approaches in terms of Hits@1. 
However, the main goal of this work is not to achieve state-of-the-art performance on standard link prediction benchmarks.
We hypothesize that this dataset is more sensitive to noisy subgraph inputs due to fewer relation types. Therefore, it is comparatively harder for the reader module to filter irrelevant context. It also leads to embedding-based approaches performing much better on this dataset.

\subsection{Comparison to other Reader Architectures} \label{sec:other_reader_comparison}
We compare our proposed Transformer-based GNN reader module with other reader model architectures for the MINERVA retriever. We use a message-passing based GNN, CompGCN \cite{vashishth2020compositionbased}, and a Transformer-based GNN, Heterogeneous Graph Transformer (HetGT) \cite{yao2020heterogeneous}, as baselines. For both datasets, our proposed Transformer-based GNN reader vastly outperforms them (Table~\ref{tab:reader_baselines}).
CompGCN uses the subgraph context to score candidate triplets. Instead of using a static context (entire KG), it now varies dynamically for each training instance. The dynamic contextualized embeddings output by the GCN encoder perform poorly with triplet-based scoring functions.
Although similar to our proposed reader, HeGT lacks some important inductive biases which are key to
a good performance --- the token type embeddings, global attention between \textit{e-e} and \textit{r-r} tokens only rather than all tokens and cross-attention. 

Given the context subgraph, a reader should learn to discard the superfluous context and reason using the relevant information to infer the correct answer. Our novel cross-attention design is a step in this direction. Expanding on this capability is a promising way to make the reader module more robust to noisy subgraph inputs.

\begin{table}[htbp]
\centering
\caption{Performance breakdown based on whether the target entity is present in the input subgraph (FB15K-237 val.\ set). The performance is significantly better when the target entity is present in the subgraph.}
\label{tab:target_ent_present}
\begin{tabular}{lcccc}
\toprule
\bfseries Retriever                 & \bfseries \specialcell{Target ent.\\coverage}        & \bfseries MRR   & \bfseries Hits@1 & \bfseries Hits@10  \\ \midrule
\multirow{2}{*}{MINERVA}     & present & \textbf{.683} & \textbf{.599} & \textbf{.857} \\
                           & absent  & .144 & .078 & .277 \\ \midrule
\multirow{2}{*}{BFS}           & present & \textbf{.626} & \textbf{.515} & .\textbf{846} \\
                           & absent  & .250 & .167 & .418 \\ \midrule 
\multirow{2}{*}{One-hop neigh.}       & present & \textbf{.949} & \textbf{.928} & \textbf{.978} \\
                            & absent  & .339 & .250 & .518 \\ \bottomrule
\end{tabular}
\end{table}

\subsection{Fine-grained Analyses} \label{sec:target_ent_present_analysis}

\noindent\textbf{Effect of Target Entity Coverage in Subgraph.} To gain further insights into the reader, we report a breakdown of the link prediction performance based on whether the target entity is present in the input subgraph (Table~\ref{tab:target_ent_present}). When the target entity is present in the subgraph, the comparative performance is very high (Hits@1 is almost $8\times$ compared to when it is absent for the MINERVA retriever). This can be explained by the fact that the coverage of the target entity provides the reader with some potentially correct reasoning paths to better establish the link between the source and the target entity.
This also gives an insight that the coverage of the target entity in the context subgraph could be a useful indicator of the retriever module's performance. \\

\begin{figure}
    \centering
    \includegraphics[width=0.8\linewidth]{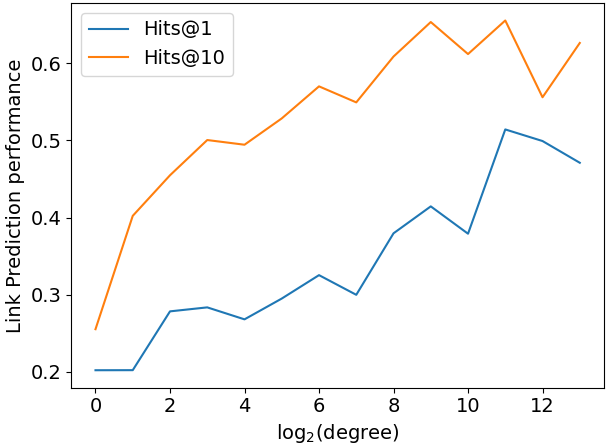}
    \caption{Link Prediction performance grouped by the logarithm of entity degree. A triple $(h, r, t)$ belongs to a degree group if either $h$ or $t$ belongs to it. The performance is significantly better for higher entity degrees.}
    \label{fig:degree_analysis}
\end{figure}

\noindent\textbf{Effect of Entity Degree.}
We analyze the effect of the degree of source and target entities in the training graph on the overall link prediction performance for FB15K-237 (Figure~\ref{fig:degree_analysis}).
We observe that the performance increases gradually with an increase in the entity degree.
Intuitively, a higher entity degree  increases the retriever's likelihood of discovering useful evidential context in the form of logical reasoning paths, connecting the source entity to the target entity for link prediction.

\begin{table*}[htbp]
\centering
\caption{Ablations for \textit{Retriever} module (validation set). The MINERVA retriever significantly outperforms other retrievers.}
\label{tab:retriever_ablations}
\begin{tabular}{cccccc|cccc}
\toprule
& & \multicolumn{4}{c}{\bfseries FB15K-237} & \multicolumn{4}{c}{\bfseries WN18RR} \\ \cmidrule{1-10}
\bfseries Model  & \bfseries Target entity coverage (\%)                                & \bfseries MRR        & \bfseries H@1    & \bfseries H@3    & \bfseries H@10   & \bfseries MRR      & \bfseries H@1   & \bfseries H@3   & \bfseries H@10   \\ \midrule
MINERVA & 46.28 & \textbf{.394}	& \textbf{.319}	& \textbf{.415}	& \textbf{.546} & \textbf{.469} & \textbf{.437} & \textbf{.476} & \textbf{.529}\\ 
BFS & 16.45 & .311	& .223	& .343	& .488 & .377	& .314	& .416	& .483\\
One-hop neigh. & .51 & .340 & .252 & .371 & .520 & .441 & .402 & .456 & .517\\ \bottomrule
\end{tabular}
\end{table*}

\begin{table*}[htbp]
\centering
\caption{Comparison of the complexity of our approach to baseline methods. Here, $n$ and $e$ denote the average number of nodes and edges in a subgraph, respectively.  $|\mathcal{E}|$ and $|\mathcal{V}|$ denote the number of edges and nodes in the KG, respectively. $T$ is the no. of iterations needed for convergence and $d$ is the hidden dimension.}
\label{tab:complexity}
\begin{tabular}{lll}
\toprule
\bfseries Model  & \bfseries Training complexity & \bfseries Inference complexity (per triplet) \\ \midrule
R-GCN \citep{schlichtkrull2018modeling} & $O(T(\mathcal{|E|}d^2))$ & $O(\mathcal{|E|}d^2)$ \\
\textbf{KG-R3} (this work)   & \specialcell{$O(T((n+e)^2 d + (n+e) d^2 +$ $d^2 + d\frac{\mathcal{|E|}}{\mathcal{|V|}}))$}  & \specialcell{$O((n+e)^2 d + (n+e) d^2 +$ $d^2 + d\frac{\mathcal{|E|}}{\mathcal{|V|}})$}\\ \bottomrule       
\end{tabular}
\end{table*}

\begin{table}[htbp]
\centering
\caption{Ablations for \textit{Reader} module (FB15K-237 val.\ set). Graph attention structure contributes the most towards final performance. Please refer to the Appendix for implementation details.}
\label{tab:reader_ablations}
\begin{tabular}{lllll}
\toprule
\bfseries Model      & \bfseries MRR   & \bfseries H@1 & \bfseries H@3 & \bfseries H@10 \\ \midrule
\textbf{Ours} & \textbf{.394} & \textbf{.319}  & \textbf{.415}  & .546   \\
\textminus \; Cross-Attention  & .387 & .312  & .410  & .541 \\ 
\textminus \; Graph attention structure & .311	& .228 & .334 & .482\\
\textminus \; Subgraph embed & .370 & .277 & .406 & \textbf{.556} \\
\textminus \; Query embed & .359 & .277 & .384 & .527\\ \bottomrule
\end{tabular}
\end{table}

\subsection{Efficiency Analysis} \label{sec:efficiency}
Table~\ref{tab:complexity} shows the comparison of training and inference complexity (per triplet) of our method with R-GCN, a prominent GNN baseline. The calculation includes the complexity of the MINERVA retriever $O(d^2 + d\frac{\mathcal{|E|}}{\mathcal{|V|}})$.
Since the subgraph size $(n+e)$ is much smaller than the number of entities in the KG (denoted by $|\mathcal{E}|$), our approach is more efficient than R-GCN both for training and inference.

\begin{table}[htbp]
\caption{Comparison of wall time of predicting a single edge for ogbl-biokg. The wall time is measured on a server with 128 CPU cores and a single A6000 GPU.}
\label{tab:inference_time}
\begin{tabular}{lc}
\toprule
\bfseries Model   & \bfseries Evaluation time per triplet (ms) \\ \midrule
CompGCN & 6.94                     \\                      
\textbf{KG-R3} (this work)    & \bfseries 1.19 \\
 \bottomrule
\end{tabular}
\end{table}
We further compare the wall time of predicting a single edge for ogbl-biokg \cite{DBLP:conf/nips/HuFZDRLCL20}. Ogbl-biokg is a large-scale biomedical KG with 93.8K nodes and 5.1M edges. Our model achieves a speedup of $5.83\times$ compared to CompGCN, a message-passing baseline (Table~\ref{tab:inference_time}). This demonstrates that our model offers a significant advantage over message-passing GNNs in terms of inference efficiency.

\subsection{Ablation Studies}

\noindent \textbf{Retriever.} For the retriever, we experiment with three choices --- MINERVA, breadth-first search, and one-hop neighborhood (Table~\ref{tab:retriever_ablations}). For both datasets, the MINERVA retriever outperforms BFS and one-hop neighborhood by a significant margin. This performance advantage can be attributed to the explicit training of MINERVA using reinforcement learning (RL) to discover paths leading to the target entity, whereas the other two approaches rely on uninformed search strategies.
To gain further insights, we analyze the statistics for target entity coverage in the subgraph (Table~\ref{tab:retriever_ablations}).\footnote{The one-hop neighborhood retriever has a low coverage as we used a small sample of one-hop edges following the HittER baseline.} As MINERVA outperforms the BFS retriever by a significant margin, this indicates that higher target entity coverage in the subgraph potentially contributes to better performance. However, the one-hop retriever outperforms BFS despite lower target entity coverage. This can be attributed to the fact that the reader better learns to ignore the noisy subgraph inputs due to the smaller context size provided by the one-hop retriever.\\

\noindent \textbf{Reader.} For the reader, we perform several ablations to evaluate the impact of different design choices. We investigate the impact of cross-attention, subgraph feature representation, query representation and using fully-connected attention in Transformer instead of the graph-induced attention structure (Table~\ref{tab:reader_ablations}).
The most significant drop in performance is caused by dropping the graph-induced attention structure, which shows that our novel attention design plays a key role in overall performance.
Among the query and subgraph feature representations, the former has a greater contribution to the performance.

\section{Conclusion and Future Work}
In this work, we propose a retrieve-and-read framework for knowledge graph link prediction. We develop a novel instantiation, KG-R3, of the framework, which consists of a novel Transformer-based graph neural network for KG link prediction. While being an initial exploration of our proposal, empirical experiments on standard benchmarks show that KG-R3 achieves competitive results with state-of-the-art methods, which indicates the great potential of the proposed framework. Our analysis offers valuable insights that can aid in the design of better retrievers within the proposed framework.
In principle, our proposed framework is scalable to large KGs, but it will require some nontrivial engineering optimizations, \eg, mixed CPU-GPU training, storing entity embedding parameters on CPU and accessing them using a push-pull API \cite{DBLP:conf/www/ZhuXTQ19, zheng2020dgl}.
Presently, the reader component, even though outperforming existing alternatives, is still prone to noisy context subgraphs.
How to further improve the robustness of the reader to noisy contexts is another major venue for future investigation.
We believe this new framework will be a valuable resource for the research community to accelerate the development of high-performance and scalable graph-based models for KG link prediction.
Future work will involve exploring the application of our framework in other knowledge graph tasks as well.

\section{Ethical Considerations}
KG link prediction aims to complete knowledge graphs by assigning higher scores to correct facts than incorrect ones. However, the top-ranked predictions might not be necessarily true, so a human verification of such predictions is required before the missing links are incorporated into the KG for public consumption. Together with human evaluation, such models have great potential to improve the coverage of KGs.

\section*{Acknowledgements}
The authors would like to thank colleagues from the
OSU NLP group and Soumya Sanyal for their valuable feedback. 
This research was supported in part by Cisco, ARL W911NF2220144, NSF OAC 2112606, NSF OAC 2118240, and Ohio Supercomputer Center \cite{OhioSupercomputerCenter1987}. 


\bibliographystyle{ACM-Reference-Format}
\balance
\bibliography{custom}

\newpage
\appendix

\setcounter{table}{0}
\renewcommand\thetable{\Alph{section}.\arabic{table}}
\setcounter{figure}{0}
\renewcommand\thefigure{\Alph{section}.\arabic{figure}}


\section{Hyperparameter and Experiment Details} \label{sec:hyper_appendix}
For the MINERVA retriever, we use decoding beam sizes of 100 and 40 for FB15K-237 and WN18RR, respectively.
The MINERVA model is trained using the default hyperparameters provided by \citet{das2018go}.
For the BFS retriever, we use up to 100 and 30 edges for FB15K-237 and WN18RR, respectively, and up to 10 outgoing edges per node. These choices are made to ensure that the subgraph size remains comparable to that of the MINERVA retriever. Following the HittER baseline \citep{chen-etal-2021-hitter}, the one-hop neighborhood retriever uses 50 and 12 edges for FB15K-237 and WN18RR, respectively.
Table~\ref{tab:hyperparam} shows the details of other hyperparameters, including the number of GPUs used, training time, and parameter count. We trained all models used in this work on Nvidia RTX A6000 GPUs. All experiments in this work correspond to a single run. 

\begin{table}[htbp]
\caption{Hyperparameter and other details for our experiments}
\label{tab:hyperparam}
\begin{tabular}{lcc}
\toprule
\bfseries Name                 & \bfseries FB15K-237 & \bfseries WN18RR \\ \midrule
Peak learning rate   & 0.01      & 0.00175  \\
No. of epochs        & 300       & 500    \\
No. of GPUs          & 2         & 1      \\
Training time (hrs.) & 67        & 18    \\
No. of parameters    & 17.3M     & 25.6M      \\
\bottomrule
\end{tabular}
\end{table}

\section{Implementation Details for Reader Ablations}\label{sec:reader_ablation_appendix}
In the ablation for omitting the query representation in the model architecture, we form the subgraph encoder input by adding the query edge \textit{([source entity], [query relation]}, \texttt{[MASK]}\textit{)} to the subgraph obtained from the retriever. Then we use the \texttt{[MASK]} token representation from the subgraph self-attention encoder for prediction. Similarly, for omitting the subgraph representation, we use just the \texttt{[CLS]} token representation from the query self-attention encoder for prediction.

\section{Implementation Details for Baseline Reader Architectures} \label{sec:other_reader_appendix}
For the CompGCN \cite{vashishth2020compositionbased} reader with the MINERVA retriever, each batch example has a different input subgraph as opposed to the entire KG in the original setup. To implement this, we train this model on 4 GPUs with batch size 1 per GPU and use gradient accumulation to simulate a batch size of 128 (as used in the original paper). We also replace batch normalization \cite{ioffe2015batch} with layer normalization \cite{ba2016layer}, as batch normalization is not suitable for batch size 1. We further reproduced the original model's performance with layer normalization. For the Heterogeneous Graph Transformer  (HetGT) \cite{yao2020heterogeneous} baseline, we use a concatenation of the subgraph feature representation and the query relation embedding as the aggregate feature representation of the model.

\end{document}